%% file: ms.tex
\documentclass[conference]{IEEEtran}
\usepackage{times}

\usepackage[numbers]{natbib}
\usepackage{multicol}
\usepackage[bookmarks=true]{hyperref}
\usepackage{graphicx}
\usepackage{epstopdf}
\usepackage{paralist}
\usepackage[linesnumbered,ruled,vlined]{algorithm2e}
\usepackage{amssymb}
\usepackage{amsmath}
\usepackage{etoolbox}
\usepackage{tikz,pgfplots}
\usepackage[export]{adjustbox}
\usepackage{caption}
\usepackage{subcaption}

\input{macros/symbols.tex}

\begin{document}

\title{A novel approach to model exploration for value function learning}


\author{\authorblockN{Zlatan Ajanovi\'{c}\authorrefmark{1},
Halil Beglerovi\'{c}\authorrefmark{2},
Bakir Lacevi\'{c}\authorrefmark{3}}
\authorblockA{\authorrefmark{1}Virtual Vehicle Research Center, Inffeldgasse 21a, 8010 Graz, Austria.}
\authorblockA{\authorrefmark{2}AVL  List  GmbH,  Hans-List-Platz  1,  8020  Graz, Austria.}
\authorblockA{\authorrefmark{3}University of Sarajevo, 71000 Sarajevo, Bosnia and Herzegovina.}
}


%

\maketitle

\begin{abstract}
Planning and Learning are complementary approaches. Planning relies on deliberative reasoning about the current state and sequence of future reachable states to solve the problem. Learning, on the other hand, is focused on improving system performance based on experience or available data. Learning to improve the performance of planning based on experience in similar, previously solved problems, is ongoing research. One approach is to learn Value function (cost-to-go) which can be used as heuristics for speeding up search-based planning. 
Existing approaches in this direction use the results of the previous search for learning the heuristics. In this work, we present a search-inspired approach of systematic model exploration for the learning of the value function which does not stop when a plan is available but rather prolongs search such that not only resulting optimal path is used but also extended region around the optimal path. This, in turn, improves both the efficiency and robustness of successive planning.
Additionally, the effect of losing admissibility by using ML heuristic is managed by bounding ML with other admissible heuristics.
\end{abstract}

\IEEEpeerreviewmaketitle

\section[\normalsize]{INTRODUCTION} 
Having fast planning algorithms is crucial for practical use of robots in changing environments and safety critical tasks.
Efficiency of Heuristic Search-based planning (A* Search \cite{hart1968formal}) largely depends on the quality of the heuristic function for estimation of the cost-to-go \cite{helmert2008how}. Ideally, if we knew the exact cost-to-go (oracle), we could find the optimal solution with minimum effort (practically traversing greedy). If the robot operates in similar environments, the learning of the cost-to-go function from previous search experience is a promising approach as Planning and Learning have complementary strengths. 
It is an issue for learning approaches to provide any guarantees on performance, have safe exploration or to learn long-term rewards. In these aspects, planning algorithms can provide valuable support. On the other hand, planning algorithms are rather slow in high dimensional spaces which can be improved if planning is properly guided.
Effective synergy  of Planning and Learning provided exceptional results so far including seminal  achievement of super-human performance in the game of Go \cite{silver2017mastering}. 

Interaction of Planning and Learning has a long history  \cite{bellman1952on}\cite{korf1990real}\cite{barto1995learning}, 
%
with several modern directions including 
End-to-end learning approximations of planning algorithms (i.e inspired by Value Iteration algorithm \cite{tamar2016value}, MCTS \cite{guez2018learning}, MPC \cite{amos2018differentiableb}),
planning to guide exploration in Reinforcement Learning \cite{weber2017imagination}, \cite{lowrey2018plan} and
learning to guide planning \cite{kim2017learning}\cite{bhardwaj2017learning} \cite{zhang2018learning}\cite{ichter2018learning}%
as well as model learning and Model Based Reinforcement Learning in general. 

This work is in direction of learning of the value function in order to guide heuristic search-based planning. It is a significantly improved extension and generalization of the work \cite{ajanovic2018safe} which was focused on automated driving application and considered the Search-Based Optimal Motion Planning framework (SBOMP) \cite{ajanovic2018search} that utilized different admissible heuristics (numeric \cite{ajanovic2017energy} and model-based \cite{ajanovic2018novel}). The admissible heuristics did not consider dynamic obstacles, so to improve performance, the authors proposed learning of heuristics which consider dynamic obstacles as well . 
The most similar approaches to the one presented in this work appeared in \cite{groshev2018learning} and in \cite{choudhury2018data}. However, in \cite{groshev2018learning} the authors used only nodes from the shortest path for learning and in \cite{choudhury2018data} the authors used backward Dijkstra's algorithm which explores the whole search space.

\begin{figure}[t]
	\centering
	\begin{subfigure}{0.22\textwidth} 
		\includegraphics[trim= 89 53 76 23,clip, width=\linewidth, frame]{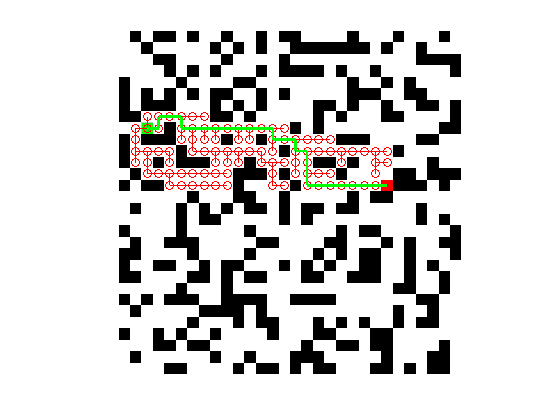}
	\end{subfigure}
	\hspace{3px} 
	\begin{subfigure}{0.22\textwidth} 
		\includegraphics[trim= 89 53 76 23,clip, width=\linewidth, frame]{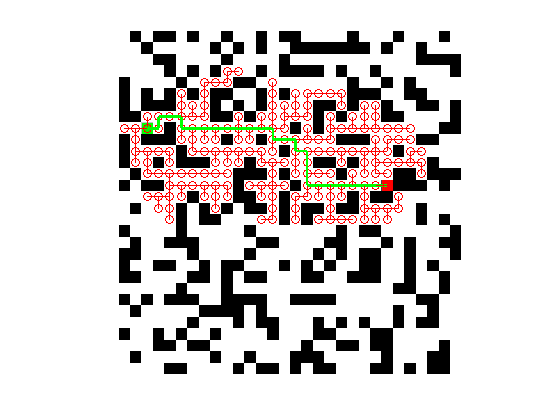}
	\end{subfigure}
	\caption{Nodes explored in Vanilla Shortest Path Problem (left) and Prolonged Heuristic Search (right).} 
	\setlength{\belowcaptionskip}{10pt}
	\label{fig:exploration}
\end{figure}

The main contribution of this work is a novel approach for efficient and systematic exploration of the models based on backward and prolonged heuristic search. This ensures that only interesting nodes are explored and all explored nodes are used for value function learning. 

\textit{Premise: For learning of the Value function it is more beneficial to explore states in the neighborhood of the optimal path (policy) than elsewhere, as the agent and the planner should spend most of the time in the neighborhood of the optimal path.}

Having explored neighboring region around optimal path helps to get back to the optimal path if the planner (or the agent) deviates and has better coverage for A* algorithm, which always looks for neighboring states.
Inspired by this idea, we prolonged the search even after the optimal path was found to explore wider region around the optimal path. Additionally, the direction of the search is reversed, such that the search starts from the goal node. In this way, all expanded nodes lead to the goal state and therefore can be used in the dataset for learning.


%
%

\section[12pt]{METHOD} 
The presented approach uses an existing admissible heuristic function ($\hadm$) and a known model $\model$ of the system to a generate dataset $\Dataset$ of exact state-cost data points. The dataset is used for supervised learning of the value function. The learned value function is then used as heuristic function $\hML$ in the search, bounded by the admissible heuristic to provide guarantees on sub-optimality.

In principle, this approach differs from reinforcement learning since it is supervised, and from imitation learning since the exact optimal solution is used instead of expert demonstrations. Nevertheless, this approach to exploration can be also used within Model Based Reinforcement Learning framework.
This approach enables theoretically inexhaustible data generation from different scenarios and initial conditions, therefore using computational resources offline to have faster planning online, when necessary. This approach can be also used in cases when it is important to focus the exploration to certain parts of the state space to reduce uncertainty in value function approximation. 

\subsection{Dataset generation}
The dataset $\Dataset$ consists of data points which carry information about the scenario (obstacles, initial and goal state) and current state together with the corresponding cost-to-go (i.e. from current state to the goal state).
For the generation of dataset $\Dataset$, planning algorithms can be used to generate data points with the exact cost-to-go.
In the vanilla Shortest Path Planning problem (SP), the goal is to find only one collision-free path (i.e. from initial to the goal state), so planning is stopped when the goal state is reached. As the goal state is reached only from one node (and each node has only one parent), the exact cost-to-go can be computed only for nodes on the optimal path. Contrary to SP, in the dataset generation, the objective is to generate as many different data points as possible. One approach is to use Backward Dynamic Programming (Dijkstra's algorithm), however this would explore the whole search space which is not practical in higher dimensional problems.


We propose a novel exploration approach based on Prolonged Heuristic Search (PHS), to generate the dataset $\Dataset$, as it is shown in Algorithm \ref{alg3}. In this approach, search is done \textit{backwards} from $\nodeG$ so that all explored nodes can be used in dataset as they contain exact cost to $\nodeG$. Additionally, as the region of higher interest is in the neighborhood of optimal path, the search is not stopped when the initial node $\nodeI$ is reached by some path (as in the SP problem), but \textit{prolonged} until $\CLOSED$ lists gets $\kEXPL$ times more nodes. This prolongation assures that more nodes in the neighborhood of the optimal path are explored.
Datapoints are constructed such that, for each node $n$ in the $\OPEN$ and $\CLOSED$ lists, the corresponding scenario structure (grid) and cost-to-go are stored in dataset $\Dataset$. Cost-to-go from node $n$ to goal node $\nodeG$ is actually cost-to-come $g(n)$ in backward search. In this way, paths do not have to be reconstructed and all expanded nodes are used in the dataset.



\begin{algorithm}[t]
\DontPrintSemicolon
\fontsize{8pt}{9pt}\selectfont
	\SetKwData{n}{$n$}
	\SetKwFunction{Expand}{Expand}
	\SetKwFunction{NewScenario}{NewScenario}
	\SetKwFunction{ProlongedHeuristicSearch}{PHS}
	\SetKwInOut{Input}{input}\SetKwInOut{Output}{output}
	\Input{$\kSCEN, \kEXPL, \model, \hadm(\cdot)$} 
	\Output{$\Dataset$\tcp*[r]{Dataset}	} 
	\BlankLine
	\Begin{
	
		$\Dataset \gets \varnothing$ \tcp*[r]{Dataset}
		\ForEach{$k \in [1, \kSCEN]$}{
 			$\langle \nodeI, \nodeG,\OBST \rangle \gets \NewScenario()$\tcp*[r]{New scenario}
%
			\tcp*[l]{Search-based Exploration}
			$\langle \OPEN, \CLOSED \rangle \gets \ProlongedHeuristicSearch (\langle \nodeI, \nodeG,\OBST \rangle, \kEXPL, \model, \hadm(\cdot) )$\;
			\tcp*[l]{Extracting data from the search}
			\ForEach{ $n \in \CLOSED  \cup \OPEN$} {	
					$\Dataset \gets \Dataset \cup (n, \OBST, g(n))$\tcp*[r]{Data points} 
					$\CLOSED \cup \OPEN \gets \CLOSED \cup \OPEN \setminus n$\;
			}
			}
		\Return{$\Dataset$}\;
		\robustify\ProlongedHeuristicSearch
	}
\caption{Prolonged Heuristic Search (PHS) for dataset generation \label{alg3}}
\end{algorithm}

\subsection{Value Function Learning} 
Learning of the Value Function $\hML$, in this approach, is a supervised learning problem (regression).
The proposed $\hML$ takes as input an image representing the current and the goal nodes (\textbf{$\nodeI$}, $\nodeG$) and a situation (obstacles $\OBST$), as can be seen in Figure \ref{fig:exploration}. (grayscale part), and returns as a result a scalar value representing an estimated cost to reach the goal from that node. 

As it is preferred that the heuristic function underestimates the exact cost (admissibility), a non-symmetric loss function can be used. Asymmetry can be introduced by augmenting Mean Square Error Loss function as:  
\begin{align}
  e_i &= y_i - \hat{y}_i, \\
  \mathcal{L} &= \frac{1}{N}  \sum_{i=1}^N  e_i^2 \cdot (\mathrm{sign}(e_i) + a)^2,
\end{align}
 with parameter  $a<0$ to emphasize the penalty for positive errors $e$.
 
\subsection{Using ML Value Function as Heuristic function}

The learned value function is used as a heuristic function in the search. To provide guarantees on sub-optimality, ML heuristic is bounded by admissible heuristic ($\hadm \leq \hML \leq \varepsilon \cdot \hadm, \quad \varepsilon \geq 1$). In this way, heuristic is $\varepsilon$-admissible so the solution is always maximum $\varepsilon$ times greater than the optimal solution \cite{likhachev2004ara}. Values of $\varepsilon$ closer to $1$ guarantee smaller deviation from optimal solution but reduce computational performance. Alternative approach would be to use Multi-Heuristic A* Search (MHA*) \cite{aine2016multi}.

\section{EXPERIMENT}

For experiment, we use grid world domain with 4-connected neighbors and 33 \% of cells are covered with obstacles in average. In total, 531 different random scenarios were used for dataset generation. From each scenario multiple data points are generated. For comparison, two datasets were created. One dataset (representing the approach from \cite{groshev2018learning}) is using only nodes from optimal path ( $\Dataset_\mathrm{VAN}$, 12.007 datapoints) and the other (as proposed in this work) uses all nodes explored in Backward Prolonged Heuristic Search ($\Dataset_\mathrm{PHS}$, 122.449 datapoints). Advantage of Backward Prolonged Heuristic Search is already visible, as from the same number of scenarios about 10 times more data points are generated, even in simple 2D problem. This is expected to be even larger in higher dimensional problems.

\subsection{Value Function approximation using Deep Learning}

In this experiment, a fully Convolutional Neural Network (CNN) was used for the value function approximation. The complete model architecture can be seen in Table \ref{tab:model}. Each layer uses the \textit{SELU} nonlinear activation \cite{klambauer2017self}. The networks were trained for $4096$ steps using a batch size of $1024$ images and a learning rate of $0.001$. The networks were initialized using the variance-scaling initializer \cite{he2015delving} and optimized with the \textit{ADAM} \cite{kingma2014adam} optimizer. In the asymmetric loss function, the parameter $a$ is set to $-2.5$.


\begin{table}[b]
\vspace{-10pt}
\renewcommand{\arraystretch}{1.3}
\centering
\caption{Model Architecture}
\begin{tabular}{r | c | c | c | c | c } 
layer & kernel & stride & dilation & avg pool & output size\\ 
\hline\hline
conv1 & $3\times3$ & $1\times1$ & $1\times1$ & none & $30\times30\times4$\\
\hline
conv2 & $3\times3$ & $1\times1$ & $2\times2$ & none & $30\times30\times8$\\
\hline
conv3 & $3\times3$ & $1\times1$ & $4\times4$ & none & $30\times30\times16$\\
\hline
conv4 & $3\times3$ & $1\times1$ & $8\times8$ & $4\times4$ & $7\times7\times32$\\
\hline
conv5 & $3\times3$ & $1\times1$ & $1\times1$ & $2\times2$ & $3\times3\times64$\\
\hline
conv6 & $3\times3$ & $1\times1$ & $1\times1$ & $2\times2$ & $1\times1\times1$\\

\end{tabular}
\label{tab:model}
\end{table} 

\subsection{Using ML Value Function as Heuristic function}

Three different heuristic functions are used in the experiment and compared based on solution quality (i.e. path length) and planning efficiency (i.e. number of explored nodes). The first heuristic function is admissible heuristic function $\hadm$ based on Manhattan distance. The second heuristic function is value function $h_\mathrm{VAN}$ trained on dataset  $\Dataset_\mathrm{VAN}$ generated from solution path only. The third function $h_\mathrm{PHS}$ is trained on $\Dataset_\mathrm{PHS}$ from the proposed Prolonged Heuristic Search.


\section{RESULTS} 

\begin{figure}[t]
	\centering
	\includegraphics[trim= 30 0 30 0,clip, width=\linewidth]{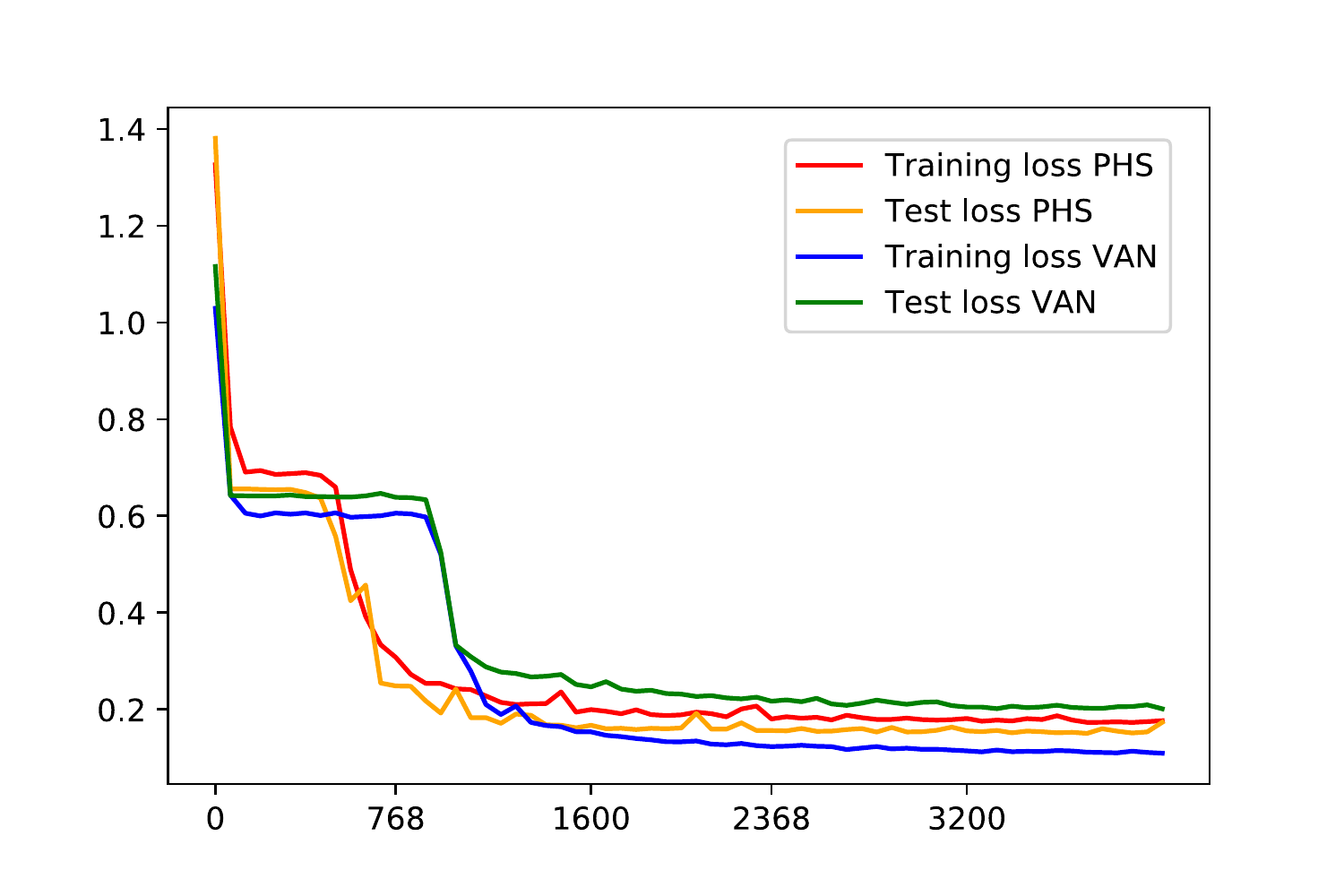}
	\caption{Training and test loss for both the $\Dataset_\mathrm{PHS}$ and $\Dataset_\mathrm{VAN}$.} 
	\label{fig:training}
\end{figure}

Experimental results support the initial premise that generating dataset using Prolonged Heuristic Search improves the quality and learnability of the value function. 
Figure \ref{fig:training} shows the training and test loss for both the $\Dataset_\mathrm{PHS}$ and $\Dataset_\mathrm{VAN}$ datasets. It can be seen that training on $\Dataset_\mathrm{VAN}$  performs poorer than training on $\Dataset_\mathrm{PHS}$, with slower convergence and a bigger difference between the training and test set, indicating overfitting. 
One reason might be the fact that $\Dataset_\mathrm{VAN}$ is smaller than $\Dataset_\mathrm{PHS}$, as the proposed approach manages to extract more data from same scenarios. The other   reason might be that the $\Dataset_\mathrm{VAN}$ was not diverse enough (i.e. many variations of the same scenario) and the network was not able to learn to generalize well. In contrast, the $\Dataset_\mathrm{PHS}$ offers more diverse training data (with many similar scenarios as all explored nodes are used), and the network is able reduce the loss much further.

\begin{figure}[t]
	\centering
	\includegraphics[trim= 0 0 0 0,clip, width=\linewidth]{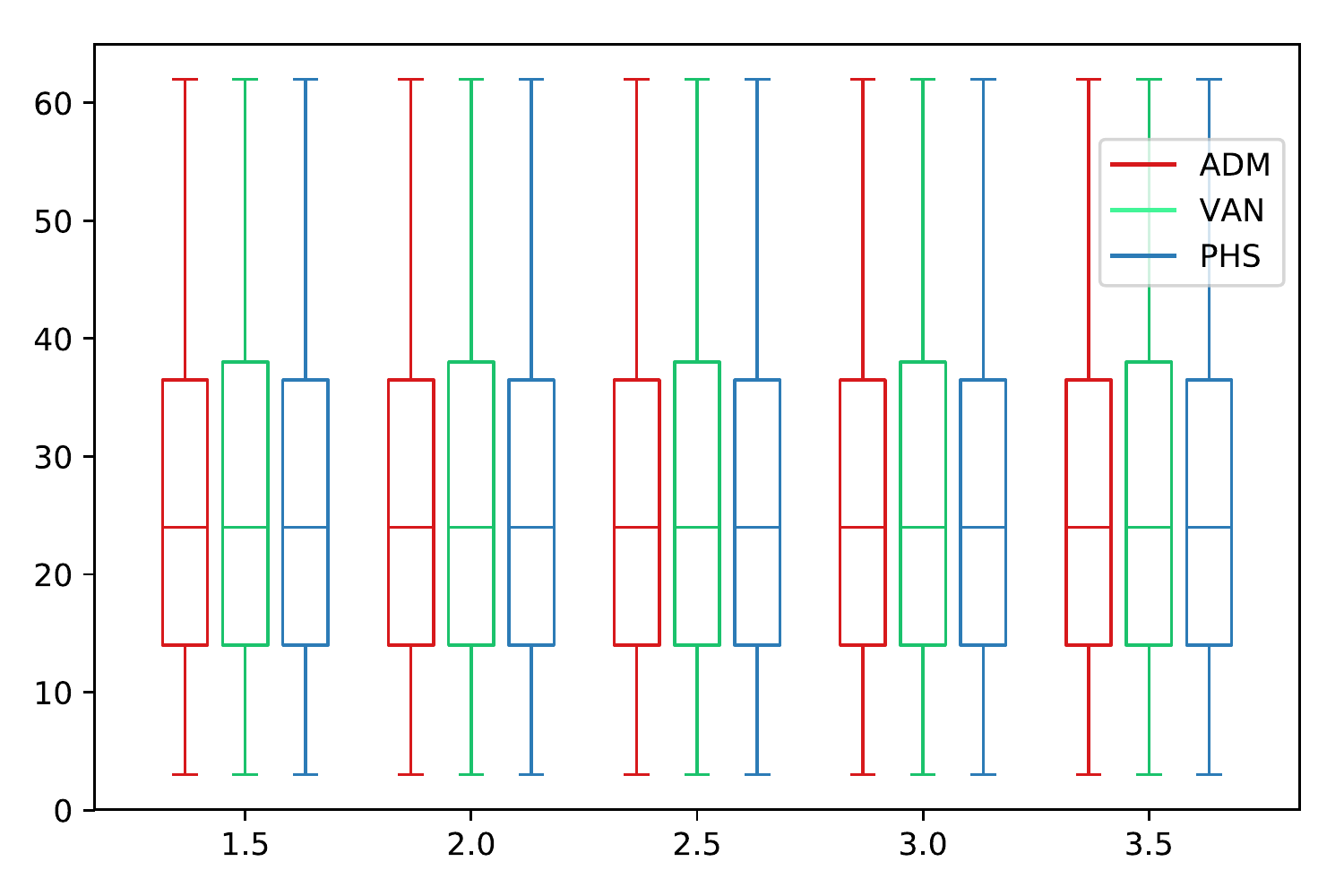}
	\caption{Path length for  $\hadm$, $h_\mathrm{VAN}$ and $h_\mathrm{PHS}$ heuristic based on $\varepsilon$ value.}  
	\label{fig:path_length}
\end{figure}

\begin{figure}[t]
	\centering
	\includegraphics[trim= 0 0 0 0,clip, width=\linewidth]{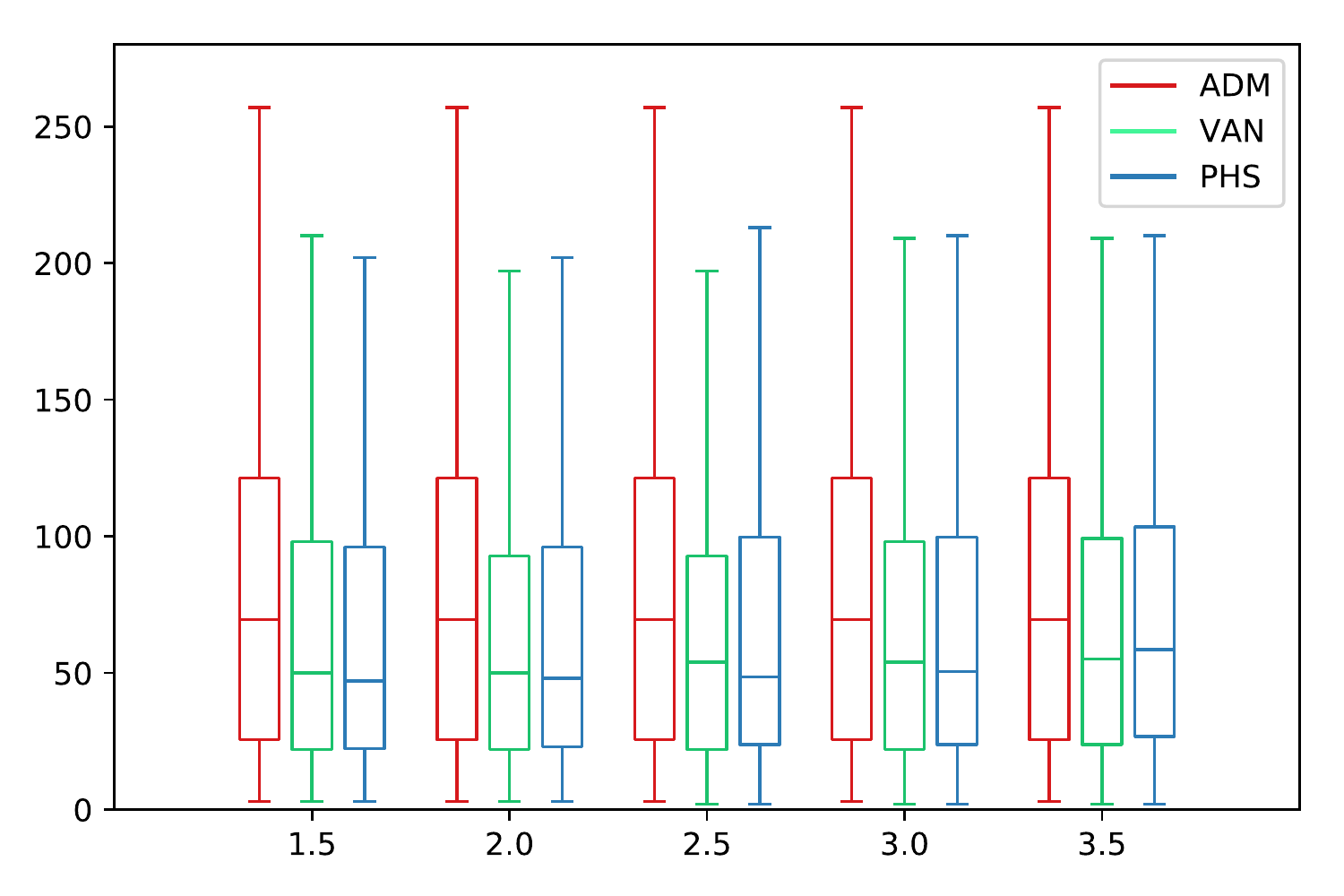}
	\caption{Number of explored nodes for  $\hadm$, $h_\mathrm{VAN}$ and $h_\mathrm{PHS}$ heuristic based on $\varepsilon$ value.} 
	\setlength{\belowcaptionskip}{10pt}
	\label{fig:exp_nodes}
\end{figure}

\begin{figure}[t]
	\centering
	\begin{subfigure}{0.24\textwidth} 
		\includegraphics[trim= 50 50 50 50,clip, width=\linewidth]{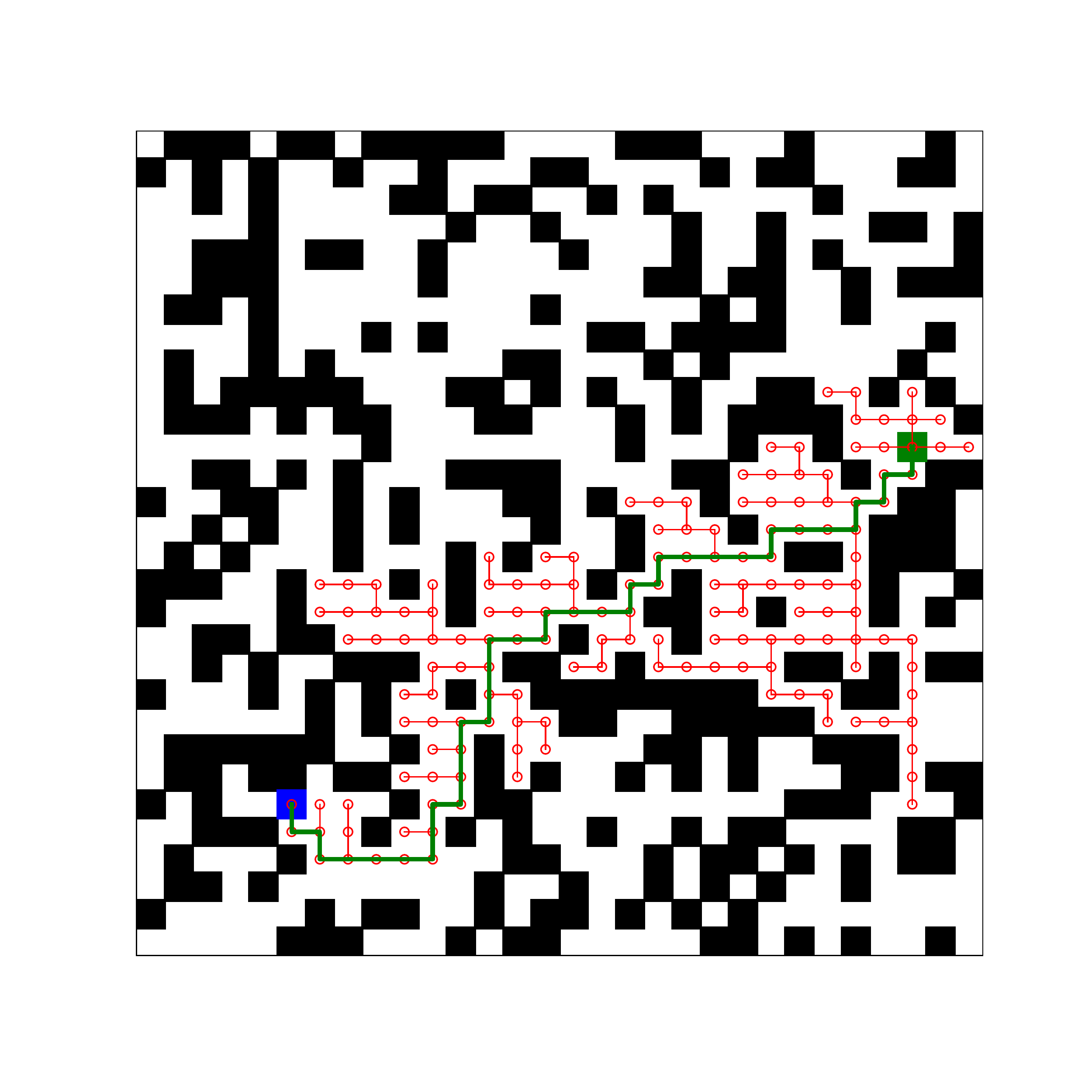}
	\end{subfigure}
	\begin{subfigure}{0.24\textwidth} 
	\includegraphics[trim= 50 50 50 50,clip, width=\linewidth]{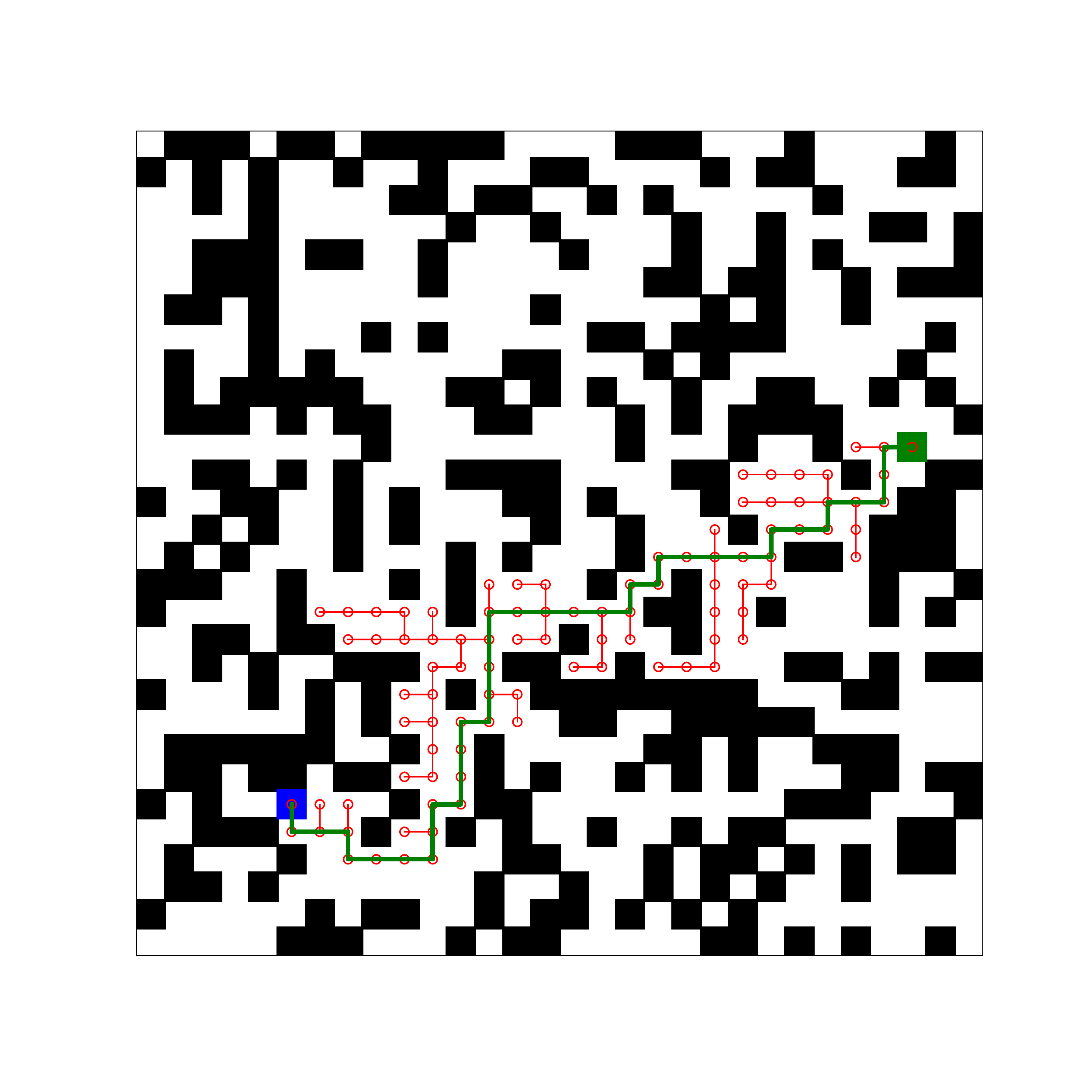}
	\end{subfigure}
	\caption{Nodes explored while planning using admissible heuristic $\hadm$ (left) and trained ML heuristic $h_\mathrm{PHS}$ (right).} 
	\label{fig:exampla}
\end{figure}

Additionally, the learned value functions were used as heuristic functions in the search. Totally $100$ random scenarios were created and both $h_\mathrm{PHS}$ and $h_\mathrm{VAN}$ were used. Results were compared based on number of explored nodes (for search efficiency) and path length (for solution quality). Both were compared with the admissible heuristic $\hadm$ while changing $\varepsilon$, which bounds the influence of ML heuristics $\hML$. Figure \ref{fig:path_length} shows that the path length does not increase significantly even for $\varepsilon = 3.5$, which means that both ML heuristics still provide solutions close to optimal. While $h_\mathrm{VAN}$ has slightly worse performance than $\hadm$, $h_\mathrm{PHS}$ has equal performance to $\hadm$ in this aspect. On the other hand, figure \ref{fig:exp_nodes} shows that both ML heuristics $h_\mathrm{PHS}$ expands less nodes than $\hadm$. In this example, performance of $h_\mathrm{VAN}$ and $h_\mathrm{PHS}$ varies slightly on $\varepsilon$, and further study on different domains and scenarios is necessary.  
Figure \ref{fig:exampla} shows an example of $h_\mathrm{PHS}$ use. It is clear from the figure that $h_\mathrm{PHS}$ expands less nodes.

%
%

\section{CONCLUSION}

The presented approach offers the possibility to effectively include Machine Learning into a deterministic planning framework, promising significant performance improvements manifested in a reduced number of explored nodes compared to those obtained using admissible heuristic ($\hadm$) while keeping guarantees on sub-optimality of the solution. The proposed approach uses the maximum of invested computational resources in planning as all expanded nodes in planning are used for learning. Experimental results showed significant improvement in search performance while keeping bounds on the sub-optimality of the solution.

Future steps would include studies of behavior in different scenarios (i.e. bug-traps and tight passages), other domains (i.e. higher dimensional problems), kinodynamic motion and extension of the approach to the end-to-end Model Based Reinforcement Learning framework.

\section*{Acknowledgments}

The project leading to this study has received funding from the European Union’s Horizon 2020 research and innovation programme under the Marie Skłodowska-Curie grant agreement No 675999, ITEAM project.\par
The study was extended in NewControl project. NewControl has been accepted for funding within the Electronic Components and Systems For European Leadership Joint Undertaking in collaboration with the European Union's H2020 Framework Programme (H2020/2014-2020) and National Authorities, under grant agreement No. 826653-2.\par
VIRTUAL VEHICLE Research Center is funded within the COMET – Competence Centers for Excellent Technologies – programme by the Austrian Federal Ministry for Transport, Innovation and Technology (BMVIT), the Federal Ministry of Science, Research and Economy (BMWFW), the Austrian Research Promotion Agency (FFG), the province of Styria and the Styrian Business Promotion Agency (SFG). The COMET programme is administrated by FFG.\par



\bibliographystyle{plainnat}
\bibliography{references}

\end{document}

%% file: macros/symbols.tex
\newcommand{\MONTH}{%
	\ifcase\month
	\or January
	\or February
	\or March
	\or April
	\or May
	\or June
	\or July
	\or August
	\or September
	\or October
	\or November
	\or December
	\fi}

\newcommand{\OBST}{\mathcal{O}}

\newcommand{\model}{\mathcal{M}}
\newcommand{\Dataset}{\mathcal{D}}

\newcommand{\nodeI}{n_{\mathrm{I}}}
\newcommand{\nodeG}{n_{\mathrm{G}}}
\newcommand{\OPEN}{\textsc{Open}}
\newcommand{\CLOSED}{\textsc{Closed}}	

\newcommand{\hadm}{h_{\mathrm{adm}}}
\newcommand{\hML}{h_{\mathrm{ML}}}
\newcommand{\ProlongedHeuristicSearch}{ProlongedHeuristicSearch}
\newcommand{\kSCEN}{k_{\mathrm{scen}}}
\newcommand{\kEXPL}{k_{\mathrm{expl}}}
